\documentclass[runningheads]{llncs}

\usepackage{eccv}

\usepackage{eccvabbrv}

\usepackage{graphicx}
\usepackage{booktabs}
\usepackage{subcaption}
\usepackage{wrapfig}
\usepackage{lipsum}

\usepackage[accsupp]{axessibility}  %

\usepackage{multirow}

\usepackage[ruled]{algorithm2e} %

\newcommand{\myparagraph}[1]{\vspace{0.1em}\noindent\textbf{#1}}
\newcommand{\larp}{LARP~}
\newcommand{\larpn}{LARP}

\newcommand\blfootnote[1]{%
  \begingroup
  \renewcommand\thefootnote{}\footnote{#1}%
  \addtocounter{footnote}{-1}%
  \endgroup
}

\usepackage[pagebackref,breaklinks,colorlinks,citecolor=eccvblue]{hyperref}

\usepackage{orcidlink}

\begin{document}

\title{Learned Neural Physics Simulation for Articulated 3D Human Pose Reconstruction} 

\titlerunning{Learned Neural Physics Simulation}

\author{Mykhaylo Andriluka\inst{1}\orcidlink{0000-0002-3109-042X} \and
Baruch Tabanpour\inst{1} \and
C. Daniel Freeman\inst{2*} \and\\
Cristian Sminchisescu\inst{1}\orcidlink{0000-0001-5256-886X}}

\authorrunning{M.~Andriluka et al.}

\institute{$^1$Google DeepMind $\qquad^2$Anthropic}

\maketitle
\begin{abstract}
We propose a novel neural network approach to model the 
dynamics of articulated human motion with contact.
Our goal is to develop a faster and more convenient alternative to traditional physics
simulators for use in computer vision tasks such as human motion reconstruction from video.
To that end we introduce a training procedure and model components that support the construction of a recurrent neural architecture to accurately learn to simulate articulated rigid body
dynamics.
Our neural architecture (LARP) supports features typically found in traditional physics simulators,
such as modeling of joint motors, variable dimensions of body parts, contact
between body parts and objects, yet it is differentiable, and an order of magnitude faster than traditional systems when
multiple simulations are run in parallel.
To demonstrate the value of our approach we use it as a drop-in replacement for a state-of-the-art classical
non-differentiable simulator in an existing video-based 3D human pose
reconstruction framework \cite{gartner2022trajectory} and show comparable or better accuracy.
\end{abstract}

\section{Introduction}
\label{sec:intro}
We\blfootnote{$^*$~Work done while the author was with Google DeepMind.} introduce a neural network approach to modeling rigid body dynamics
\cite{featherstone2007} often required for physics simulation of articulated human
motion with the objective to lower the bar for using physics-based reasoning in human
reconstruction and synthesis. 
Towards that goal, we propose a physically grounded articulated motion model that is comparable in accuracy
to state of the art classical physics simulators
(\eg \cite{coumans2015bullet,todorov2012mujoco}) but is significantly faster. 
Being developed in terms of
standard deep learning building blocks, it enables easy integration with other modern optimization and learning components.

\begin{figure}[t]
  \centering
  \includegraphics[width=0.45\columnwidth]{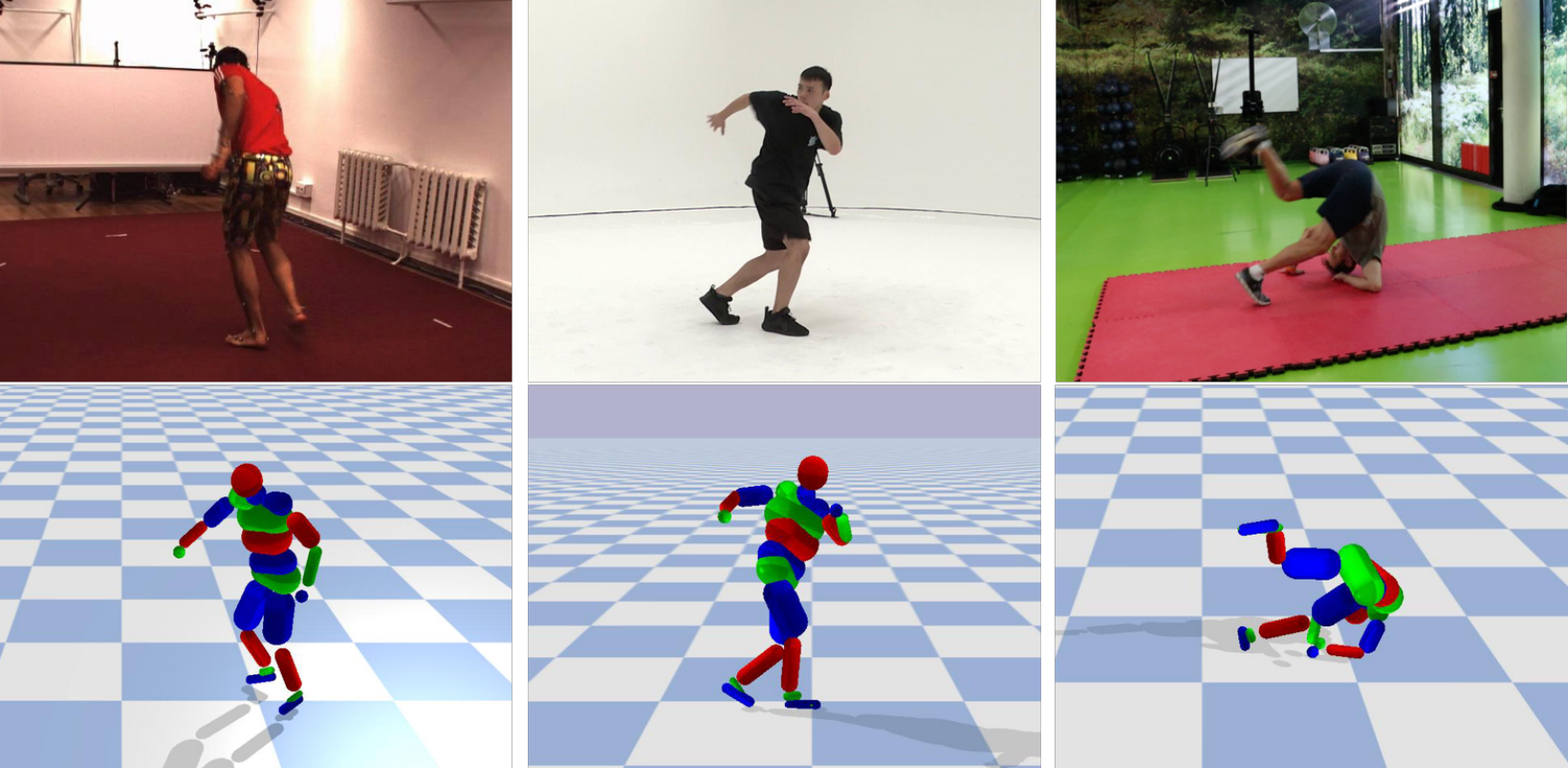}
  \includegraphics[width=0.45\linewidth]{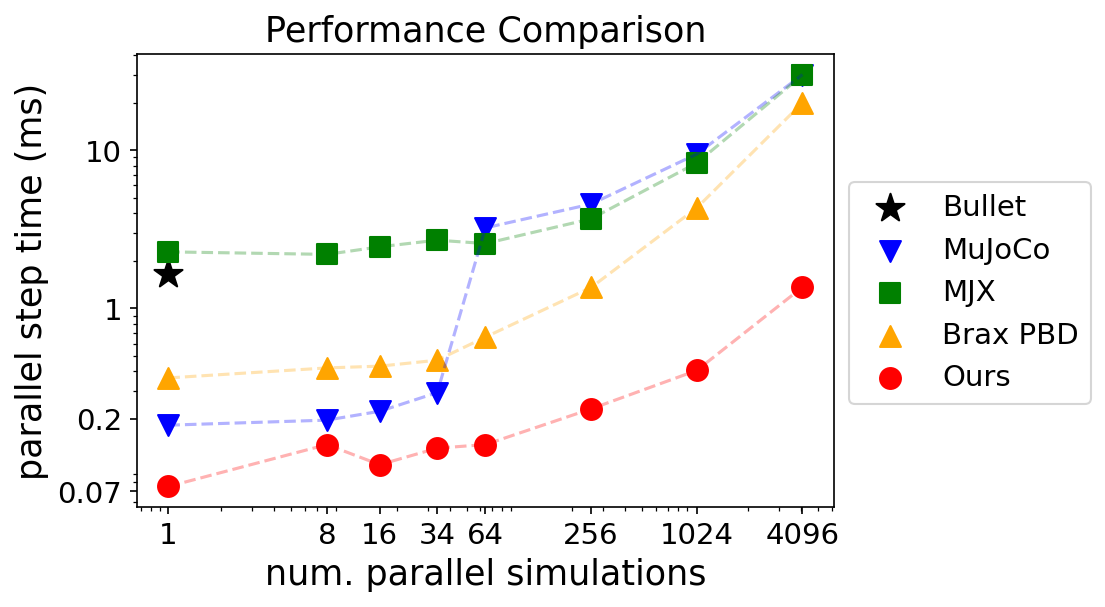}
  \caption{Left: Examples of articulated 3d human pose reconstructions obtained with \larp on public benchmarks \cite{h36m_pami,aist-dance-db} and real world video. Right: Comparison to common
    physics simulators \cite{coumans2015bullet,MJX,todorov2012mujoco,freeman2021brax} in terms of
    simulation speed.
    The x-axis shows the number of parallel simulations, whereas the y-axis shows
    the total time taken to advance all the simulations to the next step.}
  \label{fig:teaser}
  \vspace{-4mm}
\end{figure}

Our approach can be interpreted as neural simulator for a subset of rigid body dynamics and 
is closely related to an established line of work on neural simulation \cite{fussell2021supertrack,mrowca2018flexible,pmlr-v205-allen23a}.
We refer to our approach as \emph{\larp}({\bf L}earned {\bf A}rticulated {\bf R}igid body {\bf
  P}hysics).
As common in the literature, we train \larp on examples provided by a standard
off-the-shelf physics simulator, essentially learning to approximate the rigid body dynamics by means of a 
neural network.
This neural network approximation has several desirable properties. 
(1) It can be significantly faster
to execute on parallel hardware as it is composed of simple building blocks well
suited for parallelization. 
Traditional physics simulators typically solve an optimization problem at each simulation
step to compute velocities that satisfy collision and joint constraints\footnote{See Chapter
11 in \cite{featherstone2007}.}. Neural simulation can learn to approximate such calculations. In doing so it amortizes the solution of
such optimization problems at training time, and allows for faster computation at test time.
Our work is also related to how neural networks learn to directly predict the solution of optimization
problems, for \eg articulated pose estimation
\cite{kolotouros2019learning,song2020human}.  (2) Another advantage of neural simulators is their construction in terms of standard end-to-end differentiable deep learning components. 
This is in contrast to some of the established
physics simulators that are non-differentiable and not natively amenable to parallel
simulation \cite{coumans2015bullet,todorov2012mujoco}. Finally, neural physics engines can rely on parameters estimated directly from
real-world data \cite{sukhija2022arxiv,pmlr-v205-allen23a}, which enables realistic simulation even when analytic
formulations cannot be obtained.

\begin{figure}[t]
  \centering
  \includegraphics[width=0.45\columnwidth]{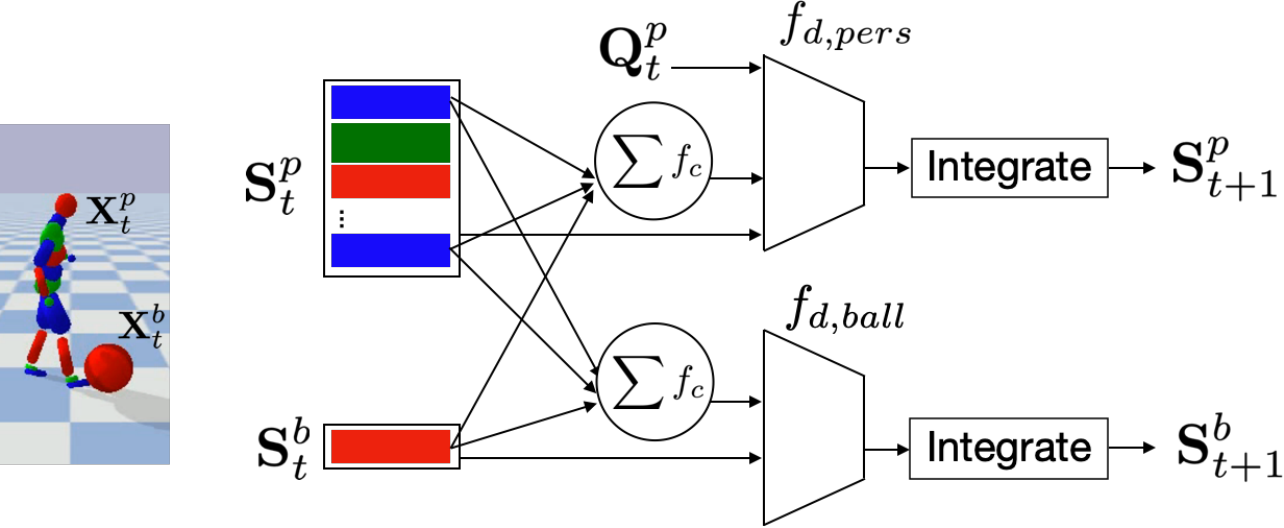}\hspace{0.7cm}
  \includegraphics[width=0.48\linewidth]{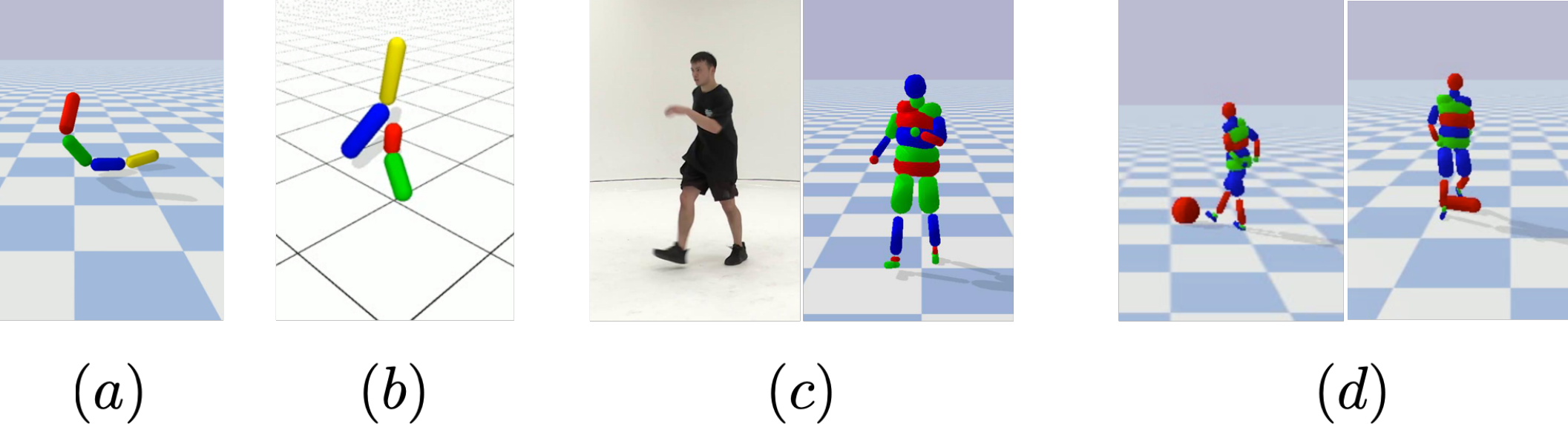} 

  \caption{Left: Overview of our approach (\larpn). At time $t$ the input to the neural simulator is
    given by a state of the scene $\mathbf{S}_t$ and joint control targets $\mathbf{Q}^p_{t}$. Here
    state is composed of the state of the person $\mathbf{S}_t^p$ and ball $\mathbf{S}_t^b$. \larp
    propagates the state through time by recurrently applying contact and dynamics networks. We
    visualize the state of each rigid component of the articulated body using rectangles with a
    color matching the scene structures. Right: Scenarios used to evaluate our approach: chain of linked capsules (a), two
    colliding capsule chains (b), articulated pose reconstruction from video (c), human-ball and human-capsule collision
    handling (d). See Supp. Mat. for videos.}
  \label{fig:overview}
  \label{fig:example_scenarios}
  \vspace{-5mm}
\end{figure}

We model the motion of a set of articulated bodies using a recurrent neural
network (RNN) with an explicit state aggregating the physical parameters of each body part. 
For each articulated object type we define a neural network 
that updates its state given the previous state, external forces and internal joint
torques applied at the current
step. Such networks, implemented as multi-layer
perceptrons (MLP), are applied recursively over time. We use different MLPs for each object, \eg person and ball in fig.~\ref{fig:overview}.
To model interactions between objects we define a collision sub-network that computes additional inputs
for the MLPs associated to each object, via sum-pooling over other objects in the scene, similarly to collision
handling in graph neural networks \cite{mrowca2018flexible}.

\myparagraph{Contributions.}  Our main contribution is a neural network architecture and training
procedure that results in a model of rigid body dynamics (\larpn) that can compute accurate human motion
trajectories up to an order of magnitude faster than traditional physics simulators (fig.~\ref{fig:teaser}). We measure accuracy both directly as well as in the context of
physics-based 3D pose reconstruction from video by integrating \larp into
the framework of \cite{gartner2022trajectory}.

Even though the
architecture of \larp is seemingly simple,
the results significantly depend
on training details such as data augmentation, gradient clipping, input features and length of sequences
used for training (\S\ref{sec:analysis-of-model}). As an additional
contribution, we perform a detailed analysis of the model components and the impact of training parameters,
identifying those ingredients that make the model perform well. We plan to make our implementation and pre-trained models publicly available upon publication.

\section{Related Work}
\label{sec:rel_work}

Our approach can be viewed as a special type of neural physics simulator focused on articulated human
motion. Neural physics is an established research area with early work
going at least back to \cite{grzeszczuk1998neuroanimator}. Neural simulation has
been applied to phenomena as diverse as weather forecasting \cite{lam23graphcast},
simulation of liquids, clothing and deformable objects \cite{cai2022active,pfaff2021learning,li2019learning}, as
well as groups of rigid bodies \cite{guo23benchmarking,allen23gnn,allen2022learning}.

Particularly relevant to our work is the literature focused on rigid body dynamics. The paradigm in this area is to formulate neural simulation as a graph-convolutional
neural network (GCN) \cite{guo23benchmarking,allen23gnn,allen2022learning}.  Recent work has shown the ability to
simulate large number of objects and correctly handle collisions even for objects with complex
shapes \cite{allen23gnn,allen2022learning}. It has been demonstrated that in certain cases
neural simulation can improve over standard rigid body simulators by directly training on real data
\cite{allen2022learning,sukhija2022arxiv}. We believe that \larp is complementary to these
works. Whereas they primarily focus on simulating large number of rigid objects passively moving
through the scene, we instead focus on modeling objects with a large number of components connected by
joints actively controlled by motors. To the best of our knowledge the only work that applies GCN to
modeling dynamics of complex articulated objects is \cite{sanchez-gonzalez2018graph}. \larp is
methodologically similar to \cite{sanchez-gonzalez2018graph}, but addresses a more complex tasks of
modeling human motion and interaction with scene objects (see fig.~\ref{fig:aist_example}).

Closely related to our work is SuperTrack
\cite{fussell2021supertrack}. As in \larpn, \cite{fussell2021supertrack} trains a fully connected
neural network that updates the dynamic state of human body parts. \larp can be seen
as a generalization of %
\cite{fussell2021supertrack}, that enables multiple objects, object collisions, improved performance
on long sequences, and support for bodies with variable dimensions of the body parts. We demonstrate
(see fig.~\ref{fig:eval_long_sequences}) that compared to SuperTrack, \larp can generate long-term motion trajectories without loss of consistency of
the simulation state\footnote{See sec.~7 and fig.~14 in \cite{fussell2021supertrack} for discussion
  of limitations.}.

\larp can be seen as an autoregressive 
motion model that can generate physically plausible human motion conditioned on the
control parameters of the body joints. The key difference to methods such as \cite{ling2020character,henter2020moglow,ghorbani2020probabilistic,habibie2017recurrent,rempe21humor} is that \larp is trained on motion trajectories generated by physics simulation and does not
include a probabilistic model for the joint control parameters. In contrast, models such as \cite{rempe21humor} are trained on motion capture data and focus on representing statistical dependencies in the articulated motion, but do not directly
incorporate physics-related motion features. We see these
approaches as complementary and hope to incorporate probabilistic motion control in the future.

\section{Methodology}
\label{sec:methodology}

\myparagraph{Overview.} 
The architecture of \larp is shown in fig.~\ref{fig:overview}. We represent the scene by a
set
of articulated objects, each corresponding to a
tree of rigid components connected by joints. We refer to such components as ``links'' using
physics simulation terminology \cite{coumans2016pybullet}. Each joint is optionally equipped
with a motor that can generate torque to actuate the object. The dynamic state of each object is
given by position, orientation, and (rotational) velocity of its links, given in
world coordinates. At each time step $t$ \larp takes a state of the scene $S_t$ and
optionally a set of motor control targets $\mathbf{Q}_t$ as input and produces the state of the
scene at the next time step. This process is applied recurrently to obtain motion
trajectories over longer time horizons.

\myparagraph{Notation.}
Let us assume that the scene is composed of $M$ articulated bodies each consisting of $N$ links. Let
$\mathbf{x}^i_{tb} \in \mathbb{R}^{3}$ denote the position of link $i \in [1 \ldots N]$ of the body
$b \in [1 \ldots M]$ at time $t$. We use $\mathbf{q}^i_{tb} \in \mathbb{R}^{4}$ for the
quaternion representing body orientation, and $\mathbf{v}^i_{tb} \in \mathbb{R}^{3}$ and
$\mathbf{\omega}^i_{tb} \in \mathbb{R}^{3}$ for linear and angular velocities, respectively.
The torque applied at time $t$ is $\mathbf{\tau}^i_{tb} \in \mathbb{R}^3$, set to
zero if no external torque was
provided. The quaternions representing the joint targets are 
$\mathbf{Q}^i_{tb} \in \mathbb{R}^4$, and are also set to zero if joint targets are not specified.
We assume that in our world representation the gravity is aligned with the z-axis, and the
ground plane corresponds to $z=0$.

\subsection{Dynamics Network}
The main component of \larp is a per-object dynamics network $f_d$. The network takes a
concatenated vector of features encoding the state of the body links, and outputs the linear and
rotational velocity of each link in the next time step. The dynamic network is implemented as a
densely connected neural network with $L=12$ layers and ELU nonlinearities
\cite{clevert2016accurate}.
In the following we describe the features used to represent state of each link. We drop the timestep
and the articulated body index to avoid clutter.

\myparagraph{Dynamic state.} The features encoding the dynamic state of the link correspond to the
root-relative position of each link $\mathbf{x}^{i}_{rrel}$, the z-position of each link
$\mathbf{x}^i_z$, world orientation of each link represented as a flattened rotation matrix
$(\mathbf{q}^i_{9d})$, and linear and angular velocities. We experimented
with other encodings of orientation such as quaternions or 6d representation and found that
 directly passing the rotation matrix works best. This representation is invariant
to shifts parallel to the ground plane which we assume orthogonal to the z-axis. 
Note that we do not encode rotation invariance
along the z-axis explicitly and instead induce it via data augmentation (\S\ref{sec:training}). We
experimentally found that taking shift and rotation invariances into account is essential in
making the approach work.

\myparagraph{Geometric information.}
For each link we include its length $l^i$, radius $r^i$ as well as the displacement error 
$\mathbf{d}^i$ of the joint to the parent link.
Our state representation independently encodes the positions of all links in world
coordinates. This is often referred to as a \textit{maximal coordinates} representation in the physics simulation
literature \cite{freeman2021brax,howelllecleach2022}.
Since positions of all links are updated independently, they might float apart and disagree on the
position of the mutual joint after a number of update steps.
To mitigate this effect we explicitly provide the difference between positions of the joint computed
based on the state of the child and parent link, and add a loss function to penalize such
displacement during training (\S\ref{sec:training}).
We experimentally show in
that these components are essential for good performance (fig.~\ref{fig:ablation-dynamics} and fig.~\ref{fig:vis_ablation_disp_loss_features}). 

\myparagraph{Control.} We include the target orientation $\mathbf{Q}^i$ of the link relative to the
parent. For traditional simulators such as \cite{coumans2016pybullet,todorov2012mujoco}
target orientation is used by the control algorithm (\eg PD-control) to compute the torque applied
by the joint motor, whereas \larp is supposed to learn from data how to transform
control targets into updates of the link state. In addition we also provide an external torque
$\mathbf{\tau}^i$ which is used in the experiments in \S\ref{sec:analysis-of-model} to diversify
the motion trajectories. 

\myparagraph{Contact.}
We include a feature vector $\mathbf{\hat{p}}^i \in \mathbb{R}^6$ computed by the
contact network $f_c$ that encodes the information about the interaction with links of other objects. The computation of the contact network $f_c$ is described in \S\ref{sec:contact}.

\subsection{Contact Network}
\label{sec:contact}
Given two articulated bodies $m$ and $b$, the contact feature $\mathbf{\hat{p}}^i_m$ is given by
\begin{equation}
\mathbf{\hat{p}}^i_m = \sum_{j}f_c(\phi^j_b, \phi^i_m, c(\phi^j_b, \phi^i_m), \theta_c).
\label{eq:contact_feat}
\end{equation}
The contact feature for link $i$ is the summation of $f_c$ over all links $j$, where $f_c$ is a neural network with $K=6$ fully-connected layers, the vector $\phi^i_b$ contains
position $\mathbf{x}^i_b$, quaternion orientation $\mathbf{q}^i_b$, velocity $\mathbf{v}^i_b$,
angular velocity $\mathbf{\omega}^i_b$, capsule length $l^i_b$, capsule radius $r^i_b$, and mass
$m^i_b$ of link $i$. The function $c(\cdot)$ computes properties of the collision between two links:
link-relative contact point, normal, and penetration distance. We use the capsule-capsule contact
implementation in Brax \cite{freeman2021brax} which is implemented in JAX and is differentiable
\cite{jax2018github}. Before feeding in features to the contact network, features are normalized by
their mean and standard deviation obtained from a single batch of offline training data. As motivated empirically in \S\ref{sec:analysis-of-model}, we apply a
stop-gradient on the contact feature before feeding it into the dynamics network to stabilize
training. Notice that the summation of $f_c$ in \eqref{eq:contact_feat} is similar in spirit to the summation to obtain net force (e.g. Netwon's Second Law), and is also similar to graph neural networks (GNN) used in neural physics simulators \cite{allen2023graph}; our approach is a special
case of GNNs with the graph between articulated objects being fully connected.

\begin{algorithm}[t]
\begin{small}
\SetAlgoNoLine
\KwIn{Dynamics network parameters $\theta_d$, contact network parameters $\theta_c$, a batch of sub-sequences $S$, dynamics network feature normalization $\mu_d$, $\sigma_d$, contact network feature normalization $\mu_c$, $\sigma_c$, dynamics network output normalization $\mu_{do}$, $\sigma_{do}$, dynamics feature function $f_x$, contact feature function $f_y$, $dt$=0.01, $M$ = 2, $N_h$ = 20 }
\KwOut{$\theta_d$, $\theta_c$}
n = 0\;
\Repeat{$n =$ Epochs}{
    $\hat{S}_0 = S_0$ \;
    \For{t = 1 $\mathbf{to}$ $N_h$ } {
        $Y_{t,1} = (f_y( \mbox{stopgrad}(\hat{S}_{t,1}) ) - \mu_c) / \sigma_c$ \\
        $Y_{t,2} = (f_y( \mbox{stopgrad}(\hat{S}_{t,2}) ) - \mu_c) / \sigma_c$\\
        $\hat{p}^{ij}_{12} = f_c(Y^i_{t,1}, Y^j_{t,2}, h(Y^i_{t,1}, Y^j_{t,2}), \theta_c)$\\
        $\hat{p}^{ji}_{21} = f_c(Y^j_{t,2}, Y^i_{t,1}, h(Y^j_{t,2}, Y^i_{t,1}), \theta_c)$\\
        $\hat{p}^i_1 = \sum_{j} \hat{p}^{ij}_{12}$\\
        $\hat{p}^j_2 = \sum_{i} \hat{p}^{ji}_{21}$\\
        \For{b = 1 $\mathbf{to}$ $M$ } {
            $X_t = (f_x(\hat{S}_t, {\hat{p}^1_b, ..., \hat{p}^N_b}) - \mu_d) / \sigma_d$\\
            $(\mathbf{\hat{v}}_{t+1}, \mathbf{\hat{\omega}}_{t+1}) = f_d(X_t, \theta_d)$ \\
            $(\mathbf{\hat{v}}_{t+1}, \mathbf{\hat{\omega}}_{t+1}) = \sigma_{do}(\mathbf{\hat{v}}_{t+1}, \mathbf{\hat{\omega}}_{t+1}) + \mu_{do}$ \\
            $\mathbf{\hat{x}}_{t+1}, \mathbf{\hat{q}}_{t+1}$ = Integrate($\hat{S}_t$, $\mathbf{\hat{v}}_{t+1}$, $\mathbf{\hat{\omega}}_{t+1}$) \\
            $\hat{S}_{t+1} = (\mathbf{\hat{x}}_{t+1}, \mathbf{\hat{q}}_{t+1}, \mathbf{\hat{v}}_{t+1}, \mathbf{\hat{\omega}}_{t+1})$
        }
    }
    Compute loss $L$ from Equation \ref{eqn:08} using $\hat{S}$ and $S$\\
    $(\theta_c, \theta_d) \leftarrow Adam(\theta_c, \theta_d, L)$
}
\caption{Algorithm for training the dynamics and contact models given two articulated bodies.}
\label{alg:one}
\end{small}
\end{algorithm}
\vspace{-0.1cm}

\subsection{Integrator}
To compute the position and orientation of all links in the next timestep, we integrate the velocity
outputs of the dynamics network $f_d$ over a timestep $dt$. The world position and
orientation of each link are given by $\mathbf{\hat{x}}_{t+1} = \mathbf{x}_t +  \mathbf{\hat{v}}_t
dt$ and $\mathbf{\hat{q}}_{t+1} = \mathbf{q}_t +   g(\mathbf{\hat{\omega}}_t, \mathbf{q}_t) dt$,
where $g(\mathbf{\hat{\omega}}_t, \mathbf{q}_t) = 0.5  quat(\mathbf{\hat{\omega}}_t) \otimes
\mathbf{q}_t$ computes quaternion derivative from rotational velocity, and
$\otimes$ is a quaternion product.

\subsection{Training}
\label{sec:training}
We estimate the parameters of the dynamics and contact networks on a dataset of training examples
corresponding to sequences of scene states comprised of the position, orientation, and 6d velocity of all links $N$ and all articulated bodies $M$, over a period of $T$ steps.
Let's denote the given a ground truth sequence of states as $S = \{S_t|t=1, \ldots, T\}$, and a sequence of states
produced by the model starting from the initial state $S_{0}$ by $\hat{S}$.
The loss used to estimate parameters consists of a position, rotation, and joint displacement loss over a rollout of length $T$. The position loss is given by
\begin{equation}
\label{eqn:06}
L_p = \frac{1}{NMT} \sum_{i, j, t} (\mathbf{x}^{ij}_t - \mathbf{\hat{x}}^{ij}_t ) ^2.
\end{equation}
The rotation loss is
\begin{equation}
\label{eqn:07}
L_r = \frac{1}{NMT} \sum_{i, j, t} 1 - | \mathbf{q}^{ij}_t \cdot \mathbf{\hat{q}}^{ij}_t |,
\end{equation}
and the joint displacement loss is the mean-squared error between the predicted child and parent joint position
\begin{equation}
\label{eqn:disploss}
L_d = \frac{1}{NMT} \sum_{i,j,t} (\mathbf{\hat{x}}^{ij}_{pt} - \mathbf{\hat{x}}^{ij}_{ct} ) ^2.
\end{equation}
The joint displacement loss encourages joint constraints to be respected. Notice that if joint constraints are not violated, $L_d=0$. The total loss is
\begin{equation}
\label{eqn:08}
L = w_p L_p + w_r L_r + w_d L_d,
\end{equation}
where $w_p=1$, $w_r=1$, and $w_d=0.1$ are hyper-parameters that balance the loss terms to have
similar weight during training. We summarize the details of  model training for the case of $M=2$ articulated bodies in alg.~\ref{alg:one}.

\myparagraph{Implementation details.}  Depending on the dataset and setting we generate state
trajectories with a length between $80$ and $200$ steps. At training time we then subdivide them
into batches of shorter subsequences of length $T$. Note that $T=1$ amounts to doing ``teacher
forcing'' on every step, meaning that the model always gets one of the ground-truth states as input.
We experimentally observe that models trained on short subsequences ($T < 5$) perform poorly (see
\S\ref{sec:analysis-of-model}). This is likely because there are subtle differences between
ground-truth states and states generated by the model, and the model needs to be exposed
to sufficiently diverse generated states at training time. To stabilize the training for larger $T$ we
add gradient norm clipping and drop training batches with the gradient norm is above the threshold
of $0.3$. 
Note that the dynamics network outputs link velocities, but the loss in eq.~\eqref{eqn:08} compares
link positions and orientations. We experimented with a loss that compares velocities directly, but
found that models trained with such loss performed poorly. 
Finally, models did not perform well unless we augmented training data by randomly rotating each batch around 
an axis orthogonal to the ground plane (z-axis in our scenes).

\section{Results}

\begin{wrapfigure}{r}{0.5\textwidth}
  \vspace{-0.6cm}
  \centering
  \includegraphics[width=0.45\columnwidth]{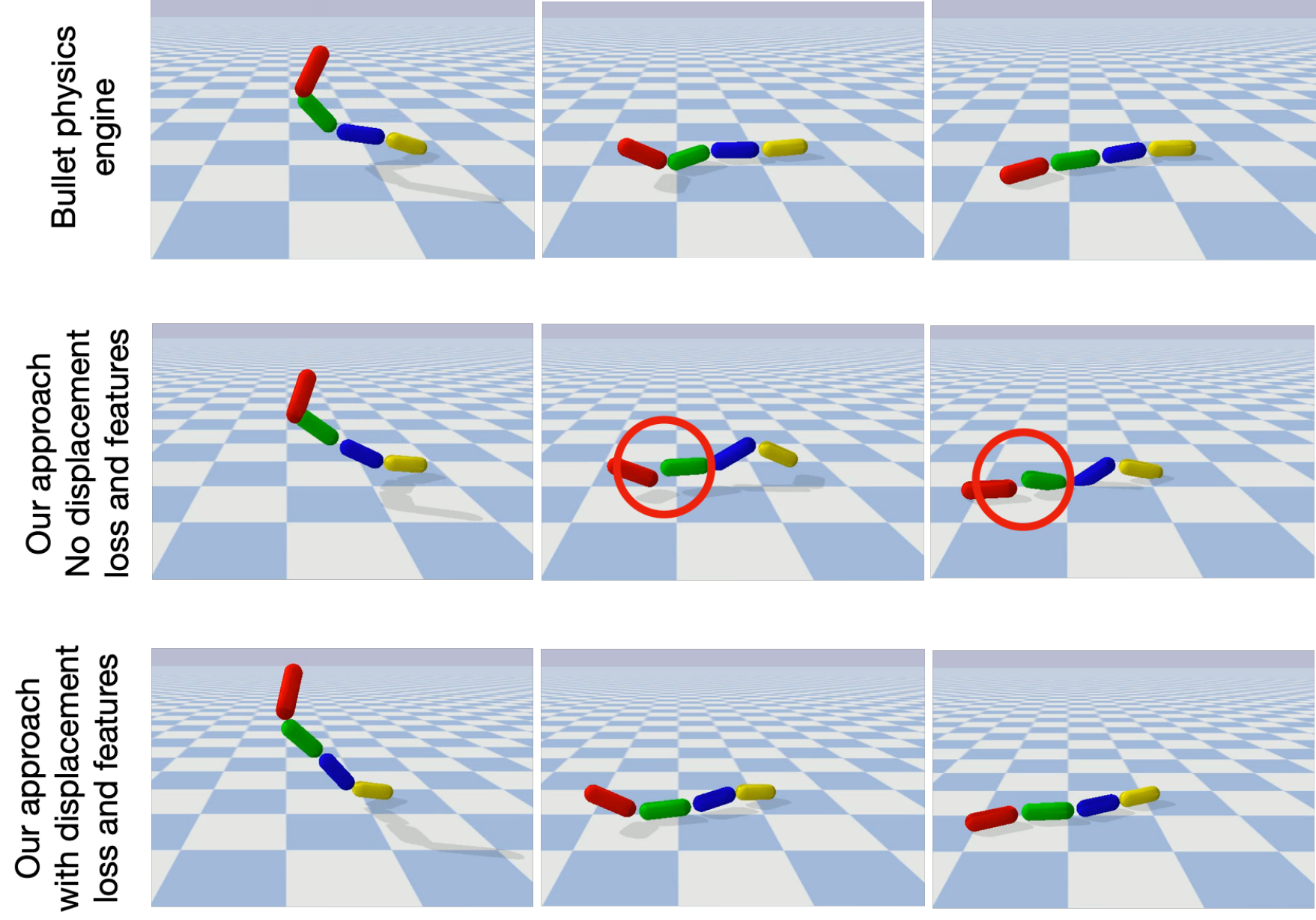}
  \caption{Example of a generated sequence obtained with a model that includes displacement features
    and displacement loss (bottom row), and without either of these elements (middle row). We
    highlight inconsistencies of the joint positions with the red circles.}
  \label{fig:vis_ablation_disp_loss_features}
  \vspace{-0.5cm}
\end{wrapfigure}

\label{sec:results}
In \S\ref{sec:analysis-of-model} we first analyze the importance of various components and the
effect of training parameters using simple articulated objects corresponding to capsule chains
(fig.~\ref{fig:example_scenarios}~(a, b)). Equipped with the best settings from this analysis, in
\S\ref{sec:results_human_video} we evaluate \larp on a more complex humanoid model with joint motors
(fig.~\ref{fig:example_scenarios}~(c, d)).
Specifically, we compare \larp to off-the-shelf physics simulator on the
task of reconstruction of human motion from monocular video. To that end
we use the physics-based articulated pose estimation approach from \cite{gartner2022trajectory} that
employs the Bullet simulator \cite{coumans2015bullet}, and replace it with \larp. 
We further evaluate the accuracy of simulating collisions between a human and an external object
represented by a ball (or capsule), quantitatively compare \larp with related work 
and benchmark \larp simulation speed against several established physics simulation engines. 

\subsection{Analysis of the model components}
\label{sec:analysis-of-model}
\begin{figure}[t]
\subcaptionbox{Subsequence window $N_h$}{
    \includegraphics[width=0.5\linewidth]{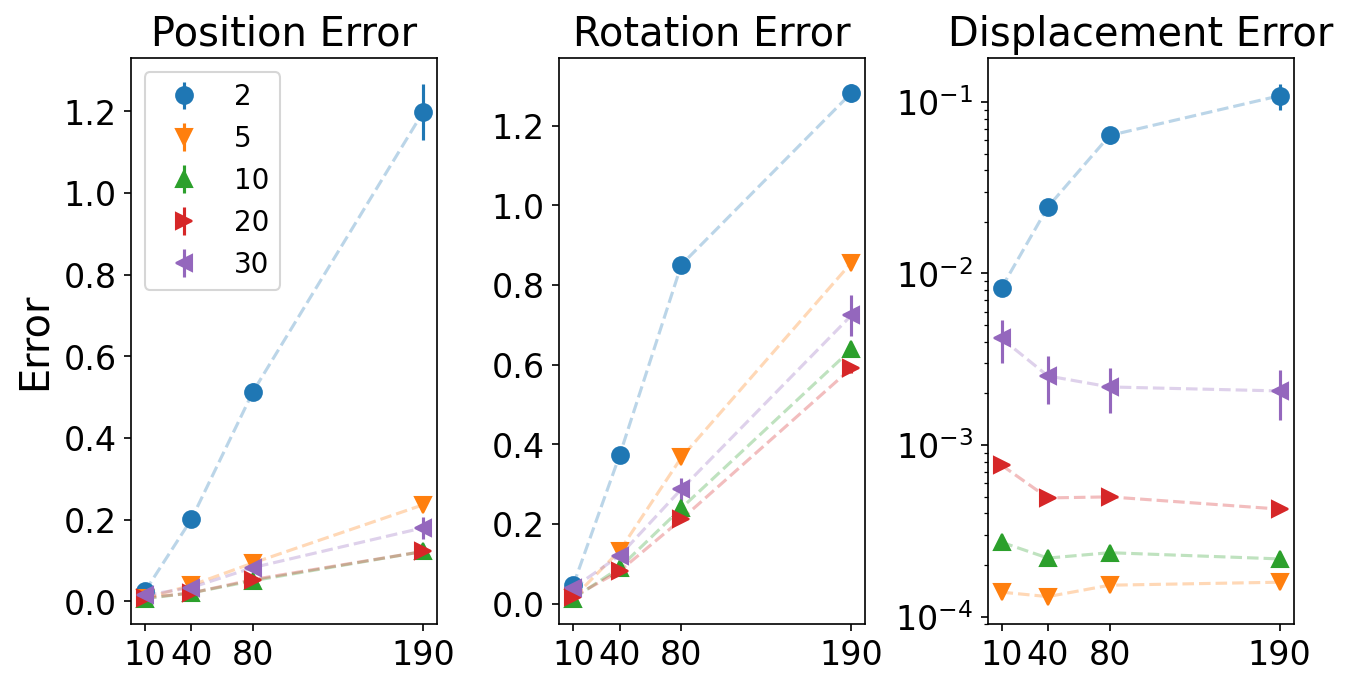}
}
\subcaptionbox{Joint Displacement}{
    \includegraphics[width=0.5\linewidth]{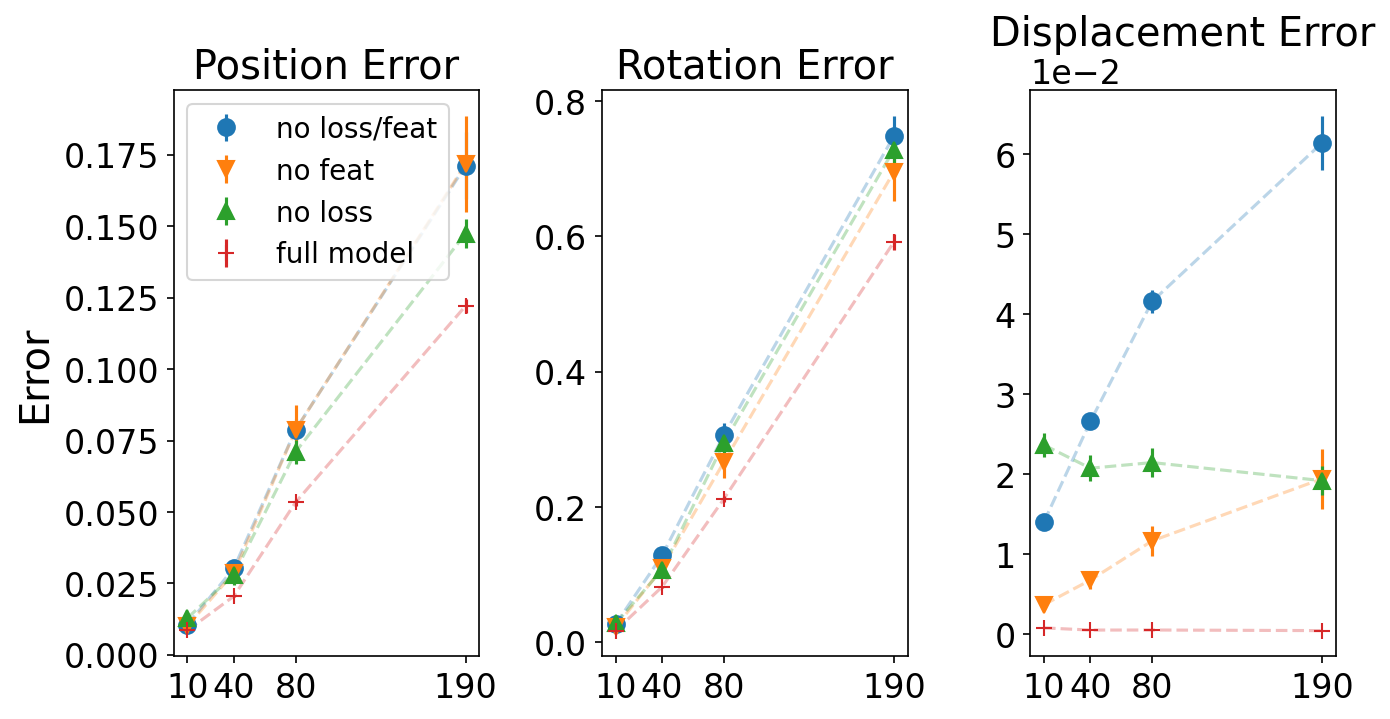}
}
\subcaptionbox{Learning Hyperparameters}{
    \includegraphics[width=0.5\linewidth]{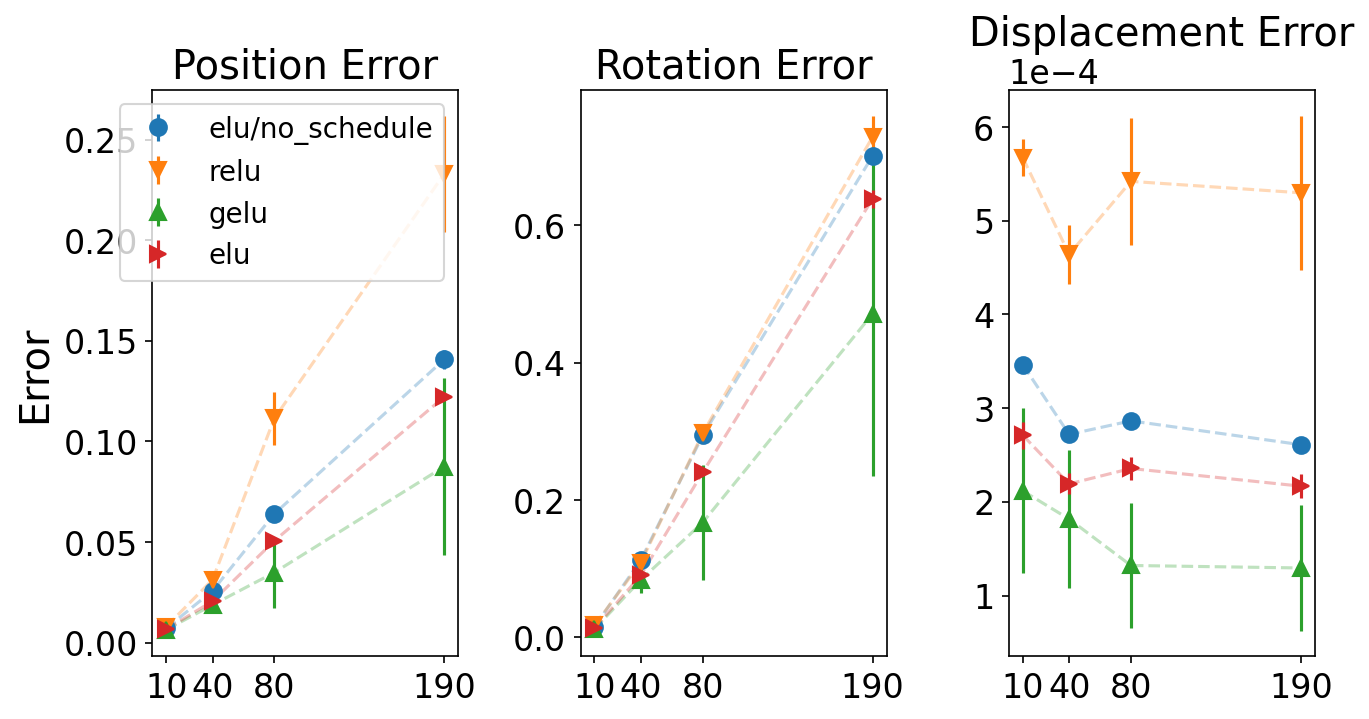}
}
\subcaptionbox{Contract network}{
\includegraphics[width=0.5\linewidth]{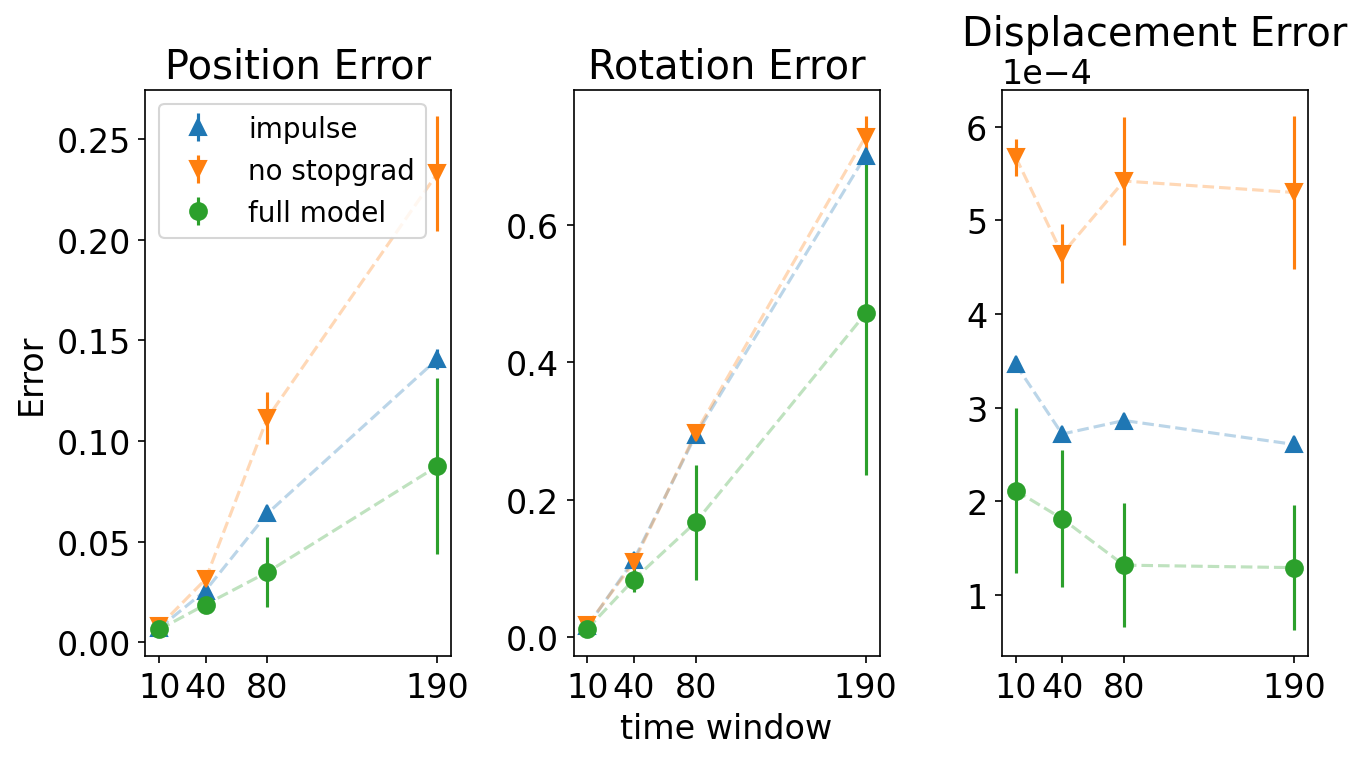}
}

\caption{Experiments with the dynamics network: training subsequence length $N_h$ (a), ablations of
  the joint displacement feature and loss (b), ablations of non-linearity type and learning rate
  schedule (c), and evaluation of variants for the contact network (d). Error bars show one standard
  deviation calculated over 5 runs. The x-axis sweeps the time window over which metrics are computed. }
\label{fig:ablation-dynamics}
\vspace{-0.3cm}
\end{figure}

\paragraph{Datasets and training.} 
We begin our evaluation using the setting shown in
fig.~\ref{fig:example_scenarios}~(a) in which four capsules linked by joints are falling on the
ground subject to random initial state and randomly applied initial torque.
The capsules can
freely rotate at the joint, and can slide and roll on the ground following impact.
This setting is simple
enough to enable quick experimentation yet is generally non-trivial since the motion of each capsule
needs to obey gravity, ground collision, and joint constraints. 
We generate 500k sequences of length 200 for training, and 25k sequences for testing and validation. Each sequence starts with a unique random
position, orientation, and velocity. We apply a random torque to the top capsule of the
chain for the first $10$ steps, and randomize capsule lengths and radii for every link.
We show an example of a generated sequence in
fig.~\ref{fig:vis_ablation_disp_loss_features} (top row).
We train the models for 200 epochs using batch size of 30k and learning rate of 1e-3. 
To reduce the effect or random initialization of neural network parameters in the following
experiments we train each network $5$ times and use the validation set to pick the best model.

\myparagraph{Analysis of the dynamics model.}

Our major finding is that a combination of training on longer sequences while using position, rotation, and 
joint displacement-based features and losses (eq.~\ref{eqn:disploss}) with applying gradient clipping, leads to
significantly improved results than reported in related work (e.g.~\cite{fussell2021supertrack}).

We observe that results improve
considerably when dynamics is unrolled for larger number of steps during training
(see~fig.~\ref{fig:ablation-dynamics}~(a)). Position error is reduced from $1.1$ meter to
$11$ cm when increasing the training sequence length from $2$ to $20$ steps. Training on
longer sequences leads to somewhat larger joint displacement error, but consistently reduces the
global position and rotation error. 
Given the setting of $N_h=20$ steps for the training sequence length we evaluate the importance of 
displacement loss and displacement features in fig.~\ref{fig:ablation-dynamics}~(b). We observe that
removing these components has a significant negative impact: position error
increases from $11$ to $15$cm, and joint displacement error increases from nearly $0$ to
$6.1$cm.
Gradient clipping was essential for training on longer sequences with $N_h > 10$. 
This is consistent with observations from training recurrent neural
networks where gradient clipping is common in stabilizing training
\cite{Goodfellow-et-al-2016}. We also found that supervising our neural simulator for positions
instead of velocities is essential for good performance. Note that our simulator outputs
velocities that are used to compute positions via integration. Replacing the position loss
(eq.~\ref{eqn:06}) with a corresponding velocity loss degrades position accuracy from $11$~cm. to $95$~cm. 
Our explanation is that the network must occasionally output small deviations from target
velocities to correct the drift in the body joints and using the velocity loss hinders such desirable behavior. 
In fig.~\ref{fig:ablation-dynamics}~(c) we verify that, as previously reported
in {\cite{fussell2021supertrack} and \cite{sukhija2022arxiv}, using soft activation
function such as ELU or GeLU is favorable compared to more standard ReLU in the context of 
neural dynamics simulation. Finally, in fig.~\ref{fig:ablation-dynamics}~(c) we show that 
learning rate schedule helps to improve accuracy.

\begin{figure}[t]
\begin{subtable}{0.45\textwidth}
  \centering
\begin{tabular}{ll|l|l|l|}
                            & $N_h$=10      & $N_h$=40      & $N_h$=80      & $N_h$=190     \\ \cline{2-5} 
\multicolumn{1}{l|}{1 link} & 0.008 & 0.021 & 0.048 & 0.173 \\ \hline
\multicolumn{1}{l|}{2 link} & 0.007 & 0.023 & 0.047 & 0.137 \\ \hline
\end{tabular}
\caption{Average difference between Bullet and \larp trajectories (in meters) for two colliding 1-link and 2-link capsule chains.\label{tab:capsule-contact}}
\end{subtable}\hfill
\begin{subtable}{0.45\textwidth}
  \centering
\begin{tabular}{lll|l|l|}
                &                              & $N_h$=10              & $N_h$=38              &
                                                                                                 $N_h$=60              \\ \cline{3-5}
   & \multicolumn{1}{l|}{Human}   & 0.010 & 0.031 & 0.048  \\
                & \multicolumn{1}{l|}{Ball}    & 0.006   & 0.033    & 0.061     \\ \cline{2-5} 
 & \multicolumn{1}{l|}{Human}   & 0.010 & 0.031 & 0.050 \\
                & \multicolumn{1}{l|}{Capsule} & 0.010  & 0.042   & 0.069     
\end{tabular}
\caption{Average difference between Bullet and \larp trajectories (in meters) for a human kicking a ball/capsule for
  sequences of $10$, $38$ and $60$ simulation steps. \label{tab:human-kick}}
\end{subtable}
\caption{Evaluation of \larp on datasets with colliding objects.}
\vspace{-0.5cm}
\end{figure}

\myparagraph{Analysis of the contact model.}
We analyze the variants of the contact network using a diagnostic environment with two chains
composed of two connected capsules shown in fig.~\ref{fig:example_scenarios}~(b).
As in experiments with the dynamics network we report position, rotation, and joint displacement
error for each variant of the contact model.
We consider the following variants: (1) our full ``contact feature'' model shown in Alg~.\ref{alg:one} where the output
of the contact network is fed as a feature to the dynamic network, (2) an alternative ``contact
impulse'' variant where the output of the contact network is used directly in the integrator as an
additive factor to linear and rotational velocities $\mathbf{v}$ and $\mathbf{\omega}$.  For
 the primary variant (1) we evaluate a version with and without stop-gradient. Results in 
fig.~\ref{fig:ablation-dynamics}~(d) indicate that all variants are able to meaningfully
 handle the contact, and that the ``contact impulse'' formulation performs somewhat worse than
 ``contact feature'' variant. Note that adding stop-gradient is important for good performance
 and that without stop-gradient the ``contact feature'' model exhibits high variance across
 retraining runs.

 We show results for the best variant on the held-out test dataset in tab.~\ref{tab:capsule-contact}.

\subsection{Simulation of articulated human motion}
\label{sec:humansimeval}
\label{sec:results_human_video}

\begin{table*}[t]
\begin{scriptsize}
\begin{center}
\begin{tabular}{c|l|c|c|c|c|c|c}
\textbf{Dataset} & \textbf{Model} & \textbf{MPJPE-G} & \textbf{MPJPE} & \textbf{MPJPE-PA} & \textbf{MPJPE-2d} & \textbf{Velocity} & \textbf{Foot skate} \\
\hline
\multirow{3}{*}{H3.6M} & DiffPhy~\cite{gartner2022diffphy} & \textbf{139} & \textbf{82} & 56 & 13 & - & 7.4  \\
& \cite{gartner2022trajectory} + Bullet& 143 & 84 & 56 & \textbf{13} & \textbf{0.24} & \textbf{4}  \\
& \cite{gartner2022trajectory} + \larp & 143 & 85 & 56 & 13 & 0.25 & 5.4 \\
\hline
\multirow{3}{*}{AIST-easy} & DiffPhy~\cite{gartner2022diffphy} & \textbf{150} & \textbf{106} & \textbf{66} & \textbf{12} & - & 19.6 \\
& \cite{gartner2022trajectory} + Bullet & 154 & 113 & 69 & 13 & 0.41 & \textbf{4} \\
& \cite{gartner2022trajectory} + \larp & \textbf{150} & 113 & 70 & 13 & \textbf{0.37} & 5 \\  
\hline
\multirow{3}{*}{AIST-hard} & \cite{gartner2022trajectory} + Bullet & 654 & \textbf{437} & 83 & 16 & 0.17 & 8.5 \\ 
& \cite{gartner2022trajectory} + \larp & \textbf{643} & 442 & \textbf{73} & \textbf{14} & \textbf{0.13} & \textbf{7.2} \\
\end{tabular}
\end{center}
\caption{Quantitative results obtained with \larp on Human3.6M~\cite{h36m_pami} and AIST
  datasets~\cite{aist-dance-db} and comparison to related work.}
\label{tab:aist}
\end{scriptsize}
\end{table*}

\begin{figure}[t]
  \centering
  \includegraphics[width=0.95\linewidth]{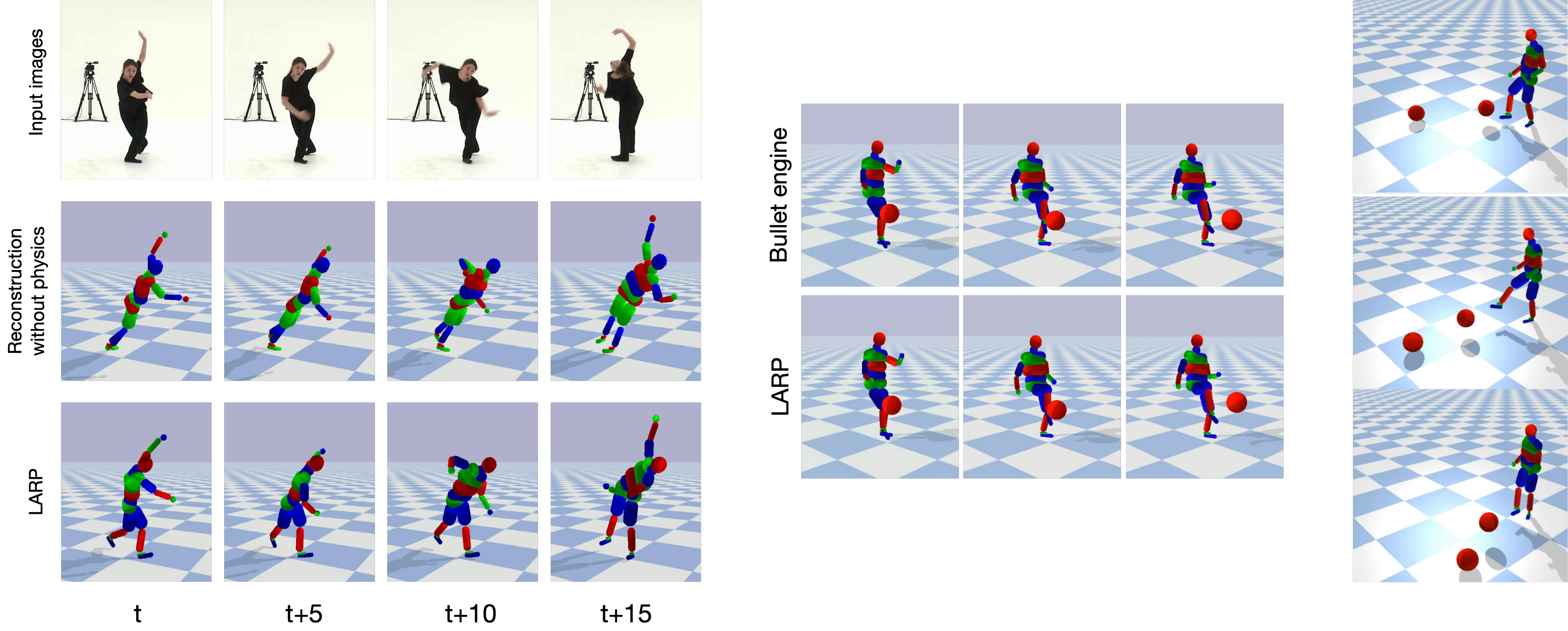}
  \caption{Left: Reconstructed 3d poses on four consecutive video frames on AIST-hard dataset. Middle row shows results
    obtained with the kinematic pipeline from \cite{gartner2022diffphy,gartner2022trajectory} that
    \larp uses for initialization. Bottom row show results obtained with \larp integrated into \cite{gartner2022trajectory}. Middle: Motion sequence with person-ball collision simulated with \larp (bottom) and comparison to Bullet engine \cite{coumans2015bullet} (top). Right: Examples of generated human motion sequences of a person kicking a ball for three different ball targets. In each image we show position of the ball right after the kick and at the end of the sequence. Note that the person pose differs considerably depending on the ball target.}
  \label{fig:aist_example}
  \label{fig:kickball}
\end{figure}

The goal of this paper is to propose a component for modeling articulated motion dynamics
suitable for variety of tasks (\eg prior for physics-based reconstruction or 
motion synthesis). We demonstrate one such application by using \larp as a
replacement for Bullet in a physics-based human motion reconstruction pipeline \cite{gartner2022trajectory}, which 
reconstructs human motion via trajectory optimization. Given an input video,
\cite{gartner2022trajectory} infers a sequence of
control parameters that results in physically simulated human motion that agrees well with 
observations (\eg locations of 2d body joints in the frames of the video input).
We refer to \cite{gartner2022trajectory} for the details and
focus on our results below.

\paragraph{Datasets.} 
Following \cite{gartner2022trajectory}
we evaluate on Human3.6M \cite{h36m_pami} and AIST\cite{aist-dance-db}. Human3.6M is a large dataset of videos of people performing common
everyday activities with ground-truth 3D poses acquired using marker-based motion capture. 
We follow the protocol proposed in \cite{PhysCapTOG2020} that excludes
``Seating'' and ``Eating'' activities. This leaves a test set of $20$ videos with $19,690$ frames in total.
AIST include diverse videos of dancing people with motions that are arguably more
complex and dynamic compared to Human3.6M. The evaluation on AIST uses pseudo-ground truth 3d joint
positions computed by triangulation from multiple camera views as defined by
\cite{li2021aistplusplus}.
We use the subset of 15 AIST videos as used for evaluation in
\cite{gartner2022trajectory,gartner2022diffphy}. 
In addition, we evaluate on $15$ AIST videos with
more challenging motions that involve fast turns and rotations.
We refer to these subsets of AIST as ``AIST-easy'' and
``AIST-hard'' in tab.~\ref{tab:aist}.

\paragraph{Model training.} 
We employ the physics-based human model introduced in \cite{gartner2022trajectory} which is based on
the GHUM model of human shape and pose \cite{xu2020ghum}. The model is composed of $26$ rigid
components represented as capsules.
We generate training data for \larp by running the
sampling-based optimization from \cite{gartner2022trajectory} on the training set and recording a
subset of human motion samples and corresponding joint torques. This produces a significantly more
diverse set of examples compared to original motion capture sequences, including examples of people
falling on the ground and various self-collisions. 
We use the best settings for \larp as identified in the experiments in
\S\ref{sec:analysis-of-model}. 

\paragraph{Metrics.}
We use standard mean per joint position (MPJPE) metrics for evaluation. The MPJPE-G metric measures
mean joint error in the world coordinates after aligning estimated and ground-truth 3d pose sequences with
respect to pelvis position in the first frame. MPJPE does pelvis alignment independently
per-frame, whereas MPJPE-PA relies on Procrustes to align both position and orientation for each
frame. Finally MPJPE-2D measures 2d joint localization accuracy. In addition we use two metrics to
measure physical motion plausibility. ``Velocity'' compares the velocity error
between the estimated motion and the ground-truth and is high when estimated motion is ``jittery''. ``Footskate'' is implemented as in \cite{gartner2022diffphy} and corresponds to the percentage of
frames where the foot moves between adjacent frames by more than $2$ cm, while in contact with the
ground. ``Footskate'' measures the presence of a common artifact in
video-based pose estimation where the positions of a person's feet unreasonably shift between nearby
frames. %

\begin{figure}[t]
  \centering
\begin{tabular}{ccc} 
\includegraphics[width=0.3\textwidth]{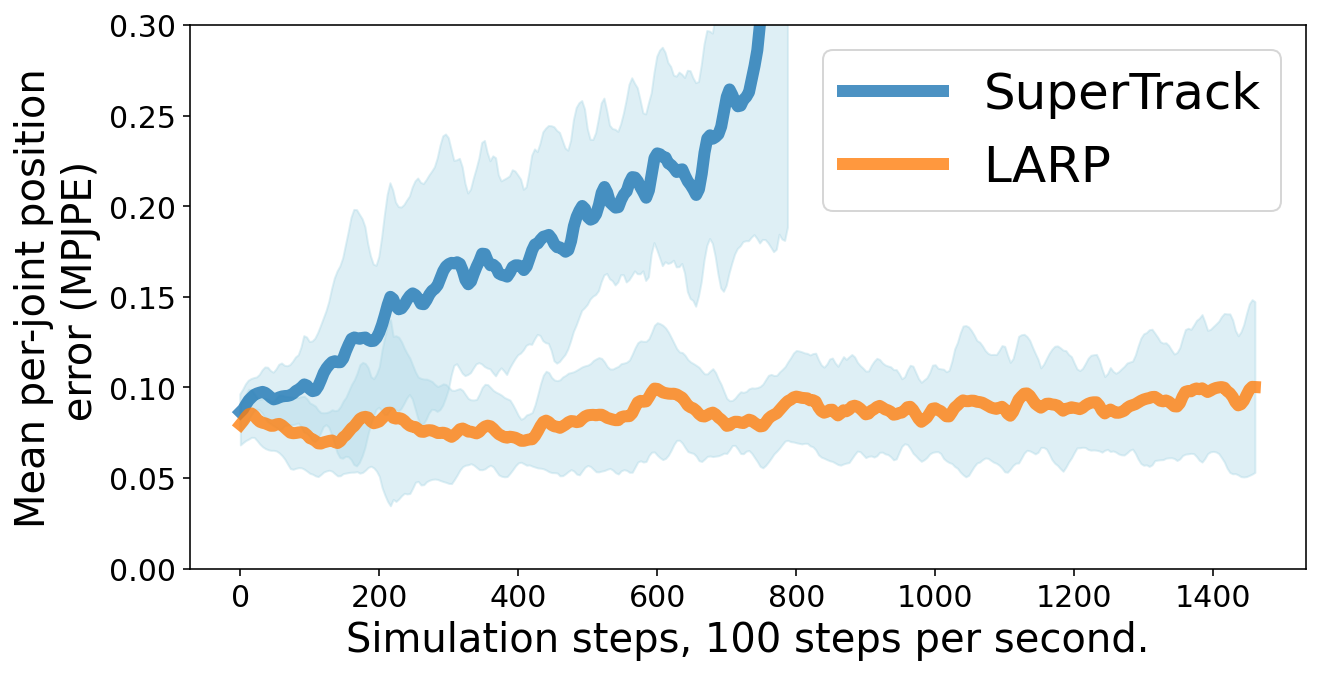} &
\includegraphics[width=0.3\textwidth]{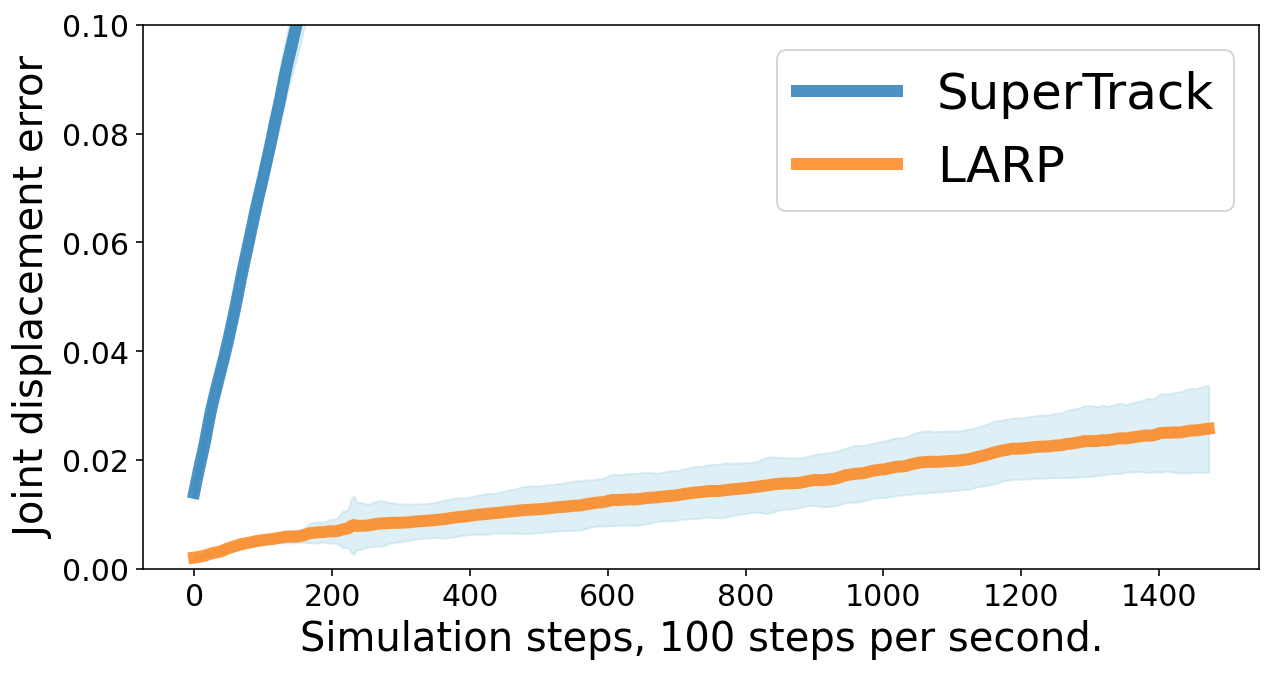} &
\includegraphics[width=0.3\textwidth]{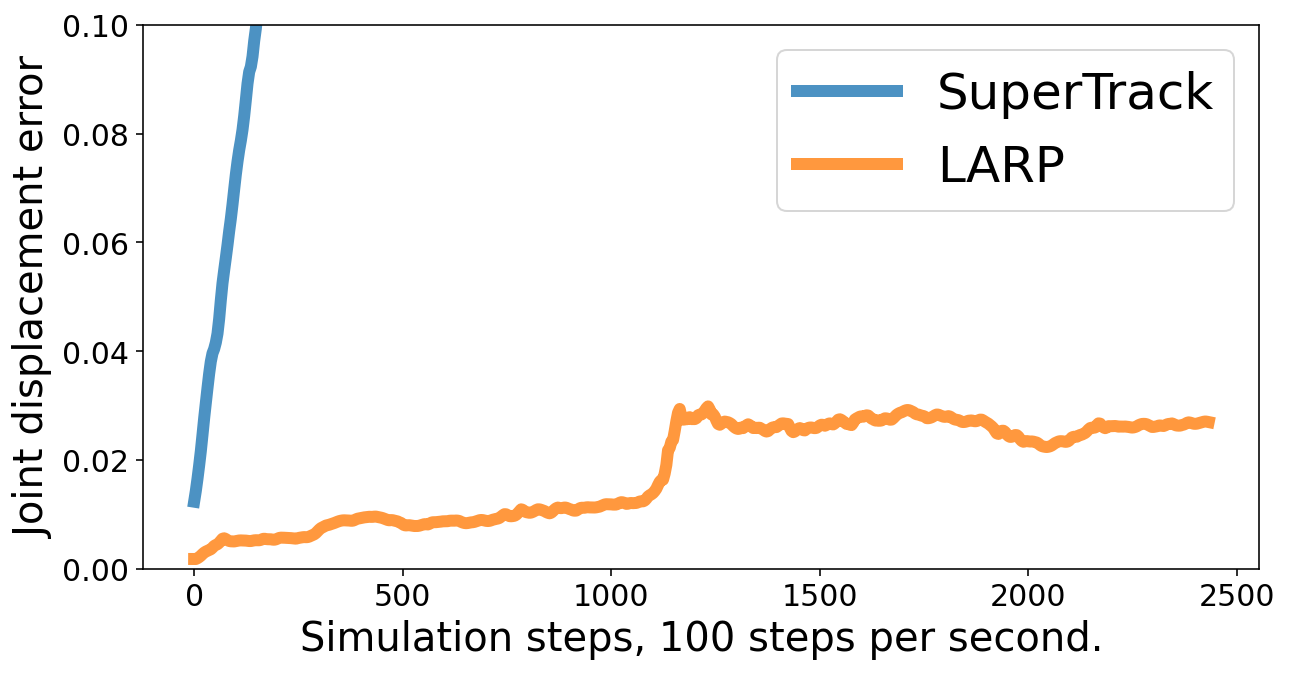}
\end{tabular}
\caption{Left and middle: average mean per-joint position error and joint displacement error
computed over all test sequences of the Human3.6M dataset. Right: displacement loss for the ``S9-WalkDog'' sequence. The units for y-axis are meters.}
  \label{fig:eval_long_sequences}
\end{figure}

\begin{figure}[t]
  \centering
  \includegraphics[width=0.95\linewidth]{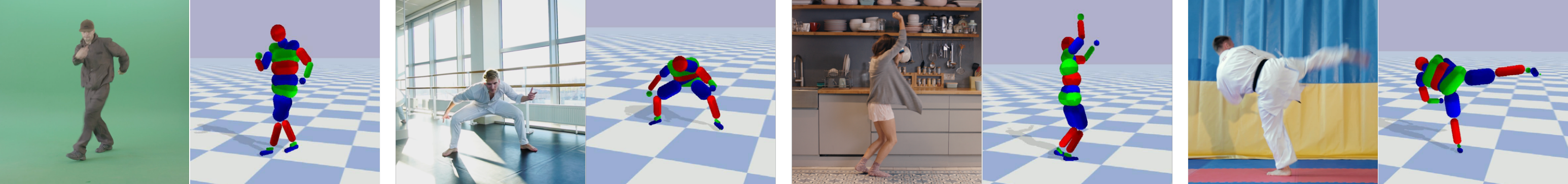}
  \caption{Example results obtained with \larp on real-world videos.}
  \label{fig:shutter}
\end{figure}

\paragraph{Results.} Our results are presented in tab.~\ref{tab:aist} and indicate that reconstructions using \larp are similar or
better compared to Bullet. For example, \larp achieves $150$ mm. MPJPE-G on AIST-easy and $643$
mm. on AIST-hard, compared to $154$ mm. and $654$ mm. for Bullet. 
Note that \cite{gartner2022trajectory} performs in general slightly worse compared to DiffPhy
\cite{gartner2022diffphy} that makes use of a differentiable physics simulator and gradient-based
optimization. In principle our approach should work in combination with \cite{gartner2022diffphy},
and we plan to explore this in the future work.  
\begin{wrapfigure}{r}{0.3\linewidth}
  \vspace{-14pt}
  \begin{center}
    \includegraphics[width=1.\linewidth]{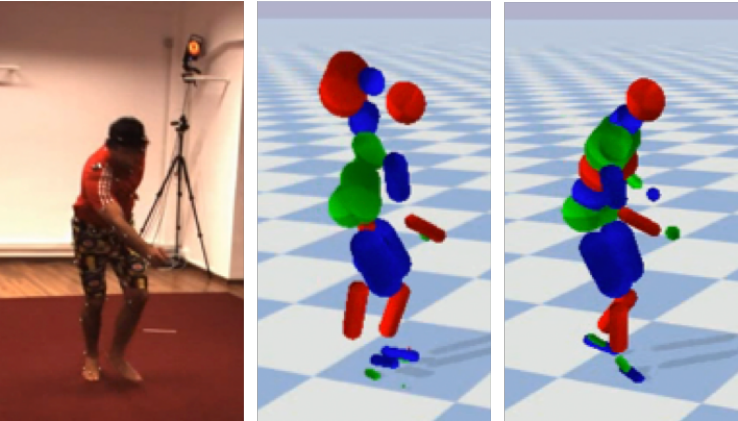}
  \end{center}
  \caption{Example of estimated pose from ``S9-WalkDog'' seq. after 11 sec. of input. Left: input frame, middle: result obtained with SuperTrack trained on longer sequences, right: result obtained with LARP.}
\label{fig:rebuttal_supertrack_example}
\vspace{-9pt}
\end{wrapfigure}

In fig.~\ref{fig:eval_long_sequences}~(left, center)
we present plots of the position error (MPJPE) and joint displacement error averaged over all test sequences of the
Human3.6m dataset\footnote{When computing the average we truncate the output to the shortest
  sequence.} and compare to the SuperTrack approach of \cite{fussell2021supertrack}\footnote{We use
  a reimplementation of SuperTrack in this comparison since the original paper didn't make the code
  or pre-trained models public.}. 
We show joint displacement error for one of the longer sequences ``S9-WalkDog''
in fig.~\ref{fig:eval_long_sequences}~(right).
We observe that for \larp the simulation indeed accumulates inconsistencies over time, albeit rather slowly,
joint displacement error grows to about 2 cm. after 1500 sim. steps. This appears to have negligible
effect on pose estimation accuracy: note that MPJPE does not increase with the simulation steps.
Our observation is that instability (e.g. inconsistent simulation state with high joint displacement
error) is not necessarily a function of the number of simulation steps, but rather happens during
complex motions that are underrepresented in the training data. For example in
fig.~\ref{fig:eval_long_sequences}~(right) the displacement error changes little for most of the 25 sec. long sequence, but jumps up during the
abrupt transition from walking into rushing forward (see fig.~\ref{fig:rebuttal_supertrack_example}). We believe that training on larger and more diverse dataset might mitigate this.

We show examples of the human pose reconstructions on AIST-easy in fig.~\ref{fig:teaser} (left plot,
middle column) and AIST-hard in fig.~\ref{fig:aist_example}~(left).  In
fig.~\ref{fig:aist_example}~(left) we compare output of \larp (bottom row) to 3d pose reconstruction
that does not employ physics-based constraints and is used by \larp for initialization (middle
row). Note that inference with \larp was able to correct physically implausible leaning of the
person and estimate correct pose of the lower body. Finally, in fig.~\ref{fig:shutter} and
fig.~\ref{fig:teaser} (left, third column) we include a few examples of 3d pose reconstructions
obtained with \larp on real-world videos. These results confirm that \larp generalizes beyond AIST
and Human3.6M datasets, is able to handle motions different from the
training set (e.g. karate kick in fig.~\ref{fig:shutter}) and can handle complex contact with the
ground (e.g. safety roll in fig.~\ref{fig:teaser}).

\myparagraph{Simulation speed.}
\label{sec:results_sim_speed}
We measure the speed of our approach using humanoid environment in Fig.~\ref{fig:example_scenarios}
(c) and compare it to Bullet \cite{coumans2016pybullet}, MuJoCo \cite{todorov2012mujoco}, and Brax \cite{freeman2021brax}. 
Bullet and $\mbox{MuJoCo}$ were run on an AMD 48-core CPU for best performance. Brax, MuJoCo XLA (MJX) \cite{MJX},
and LARP are implemented to run on SIMD architecture and are thus run on a Tesla V100
Nvidia GPU for best performance\footnote{Brax PBD is an implementation of
position based dynamics \cite{macklin2016xpbd} which is unstable for the given mass/inertia in the
humanoid model, but is added for speed comparison.}. 
For each simulator we measure the time required to perform a simulation step for different number of
simulations run in parallel. 
We run open-loop rollouts of 100 steps averaged over 5
runs for each number of parallel simulation. The compilation time is excluded. 
The results are shown in fig.~\ref{fig:teaser}. 
Overall, we observe that \larp 
exhibits better performance already when performing a single simulation ($0.07$ms for \larp
vs.~$0.19$ms for MuJoCo and $1.8$ ms. for Bullet). When running multiple simulations in parallel we improve over other simulators by an order of magnitude, \eg $1.3$ms for
$4,096$ parallel simulations compared to about $20$ ms for MJX and MuJoCo.

\myparagraph{Human-object collision handling.}
\label{sec:results_human_object}
We include two types of experiments to evaluate accuracy of human object collisions.
In the first experiment a ball or a capsule is added in a
random position and orientation in front of the humanoid. We then run simulation in Bullet engine
\cite{coumans2015bullet} and in \larp and measure the difference between trajectories of the ball
and humanoid. We show an example of simulation results in fig.~\ref{fig:kickball} (middle) and
present quantitative evaluation in tab.~\ref{tab:human-kick}. Overall we observe that \larp
fairly accurately approximates the output of a Bullet simulation. For example, the difference in the
ball position after $60$ simulation steps is about $6$~cm. on average.
In the second experiment we demonstrate that we can use \larp to control a humanoid in order to kick
the ball to a particular target. The control parameters are inferred via model predictive control by
minimizing residuals of: (1) the distance between the ball trajectory and the target, and (2) a term
that keeps the resulting articulated human motion close to the reference kicking motion. We obtain
an average error between the ball and the target of $5.7$ cm. The qualitative results are shown in
fig.~\ref{fig:kickball} (right).

\section{Conclusion}

We have presented a methodology  (\larpn) to train neural networks that can simulate the complex physical motion of an articulated human body. \larp supports features found in classical rigid body dynamics simulators, such as joint motors, variable dimensions of the body component volumes, and contact between body parts or objects. Our experiments demonstrate that neural physics predictions produce results comparable to traditional simulation, while being considerably simpler architecturally, comparable in accuracy, and much more efficient computationally. 
Our neural modeling replaces the complex, computationally expensive operations in traditional physics simulators with efficient forward state propagation in recurrent neural networks.
We discuss and illustrate the capability of \larp  in challenging scenarios involving reconstruction of human motion from video and collisions of articulated bodies, with promising results.

\bibliographystyle{splncs04}
\bibliography{biblio}
\end{document}